\PassOptionsToPackage{table}{xcolor}
\documentclass{article} 
\usepackage{iclr2027_conference,times}

\usepackage{amsmath,amsfonts,bm}

\def\eqref#1{equation~\ref{#1}}

\def\1{\bm{1}}

\DeclareMathAlphabet{\mathsfit}{\encodingdefault}{\sfdefault}{m}{sl}
\SetMathAlphabet{\mathsfit}{bold}{\encodingdefault}{\sfdefault}{bx}{n}

\usepackage{hyperref}
\usepackage{url}

\usepackage{graphicx}
\usepackage{subcaption}
\usepackage{capt-of}

\usepackage{enumitem}
\usepackage{pifont}
\usepackage{booktabs}
\usepackage{multirow}
\usepackage{makecell}
\usepackage{arydshln}
\usepackage{amssymb}
\usepackage{amsmath}
\usepackage{tcolorbox}
\tcbuselibrary{skins}
\newtcolorbox{promptbox}[1][]{
  colback=gray!5,
  colframe=black!40,
  colbacktitle=black!55,
  coltitle=white,
  fonttitle=\bfseries,
  title=#1,
  enhanced,
  boxrule=0.6pt,
  arc=2pt,
  left=6pt,
  right=6pt,
  top=6pt,
  bottom=6pt
}

\usepackage{xcolor}
\usepackage{soul}
\definecolor{my_color}{RGB}{220,240,220}

\title{When Words Fall Short: Iterative Synergy Between Verbalized Reasoning and Hidden Features for LLM Confidence Estimation}

\author{Yekun Xu$^{1,2}$ \; Ante Wang$^{2}$ \; Jingyi Ren$^{2,3}$ \; Xuanyi Chen$^{3}$ \; {\bf Weizhi Ma}$^2$\thanks{Corresponding authors.} \;\;\; {\bf Yang Liu}$^1$$^,$$^2$$^,$$^3{^*}$ \\
      $^1$College of AI, Tsinghua University, Beijing, China \\ 
      $^2$Institute for AI Industry Research (AIR), Tsinghua University, Beijing, China \\
      $^3$Dept. of Comp. Sci. \& Tech., Institute for AI, Tsinghua University, Beijing, China \\
      \texttt{\{mawz, liuyang2011\}@tsinghua.edu.cn}
}

\iclrfinalcopy 
\begin{document}

\maketitle

\begin{abstract}
Confidence estimation is crucial for developing trustworthy large language models (LLMs), with most methods following estimator-based or verbalization-based paradigms. While recent research increasingly focuses on improving verbalized self-reports of confidence, we challenge the prevailing 
view that this approach surpasses independent confidence estimators. Our empirical study shows that a dedicated confidence estimator can substantially outperform verbalized confidence, indicating that LLMs' internal representations contain richer confidence signals. Building on this finding, we propose \textbf{I}terative \textbf{Po}licy-\textbf{E}stimator \textbf{T}raining (\textbf{IPoET}), a framework that synergizes the complementary strengths of verbalized reasoning traces and informative representations. IPoET alternates policy optimization with estimator updating, integrating estimator-derived confidence feedback into policy learning and refreshing the estimator on 
new policy rollouts. Experiments across diverse datasets and Qwen and Llama backbones demonstrate that, by iteratively exploiting richer hidden features and adapting to the evolving policy distribution, IPoET consistently outperforms both estimator- and verbalization-based baselines in-domain and achieves superior or comparable results across all out-of-domain metrics. For more details, refer to \url{https://github.com/xyk829/ipoet}.
\end{abstract}

\section{Introduction}

Large language models (LLMs) have become increasingly capable across a wide range of challenging tasks, such as reasoning, code generation, and planning~\citep{jaech2024openai, guo2025deepseek, jimenez2024swe, xiao2024flowbench}. However, higher accuracy does not necessarily mean that models can correctly assess whether their outputs should be trusted. LLMs frequently generate hallucinated or incorrect answers with unwarranted confidence~\citep{xiong2024can, anh2025survey}. This lack of reliable confidence estimation severely limits the practical deployment of LLMs in high-stakes domains such as medicine, law, and finance, where decision-making errors carry significant consequences~\citep{li2024agent, siino2025exploring, li2023large, kumaran2026competing}.

Existing research on confidence estimation can generally be categorized into two paradigms:
(i) \textit{Estimator-based methods}. These methods train independent models or lightweight probing heads to exploit features embedded within the model's hidden states~\citep{beigi2024internalinspector, mahaut2024factual}.
(ii) \textit{Verbalization-based methods}. This paradigm leverages the linguistic reasoning of LLMs and derives verbalized confidence scores through prompting or training~\citep{yang2024verbalized, li2026conftuner}. Recent work on this task has increasingly focused on RL-based verbalization methods, with several studies reporting that verbalized confidence outperforms estimator-based methods~\citep{damani2025beyond, bani2025rewarding, ma2026decoupling, zhang2026confidence}.

\begin{figure}[t]
    \centering
    \captionsetup[subfigure]{skip=1pt}
    \begin{subfigure}[t]{0.54\linewidth}
        \centering
        \includegraphics[width=0.95\linewidth]{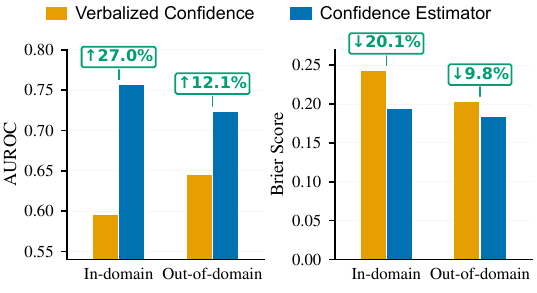}
        \caption{}
        \label{fig:header}
    \end{subfigure}
    \hfill
    \begin{subfigure}[t]{0.45\linewidth}
        \centering
        \includegraphics[width=0.95\linewidth]{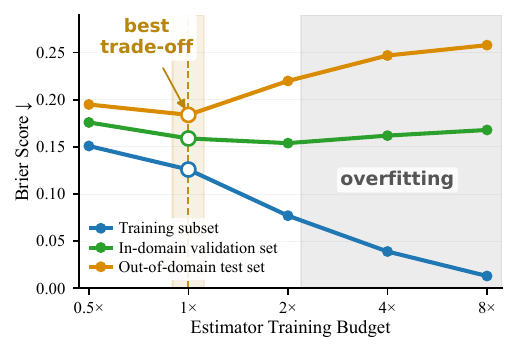}
        \caption{}
        \label{fig:training-sensitivity}
    \end{subfigure}

    \caption{
    (a) In-domain and out-of-domain comparison of confidence estimator and verbalized confidence using AUROC and Brier score. Percentages denote relative improvements over verbalized confidence.
    (b) Estimator performance across training budgets: Brier scores on a training subset, an in-domain validation set, and an out-of-domain test set illustrate overfitting with extended training.
    }
    \label{fig:empirical-study}
\end{figure}

However, revisiting this comparison leads us to a different finding: with validation-based checkpoint selection, an independent confidence estimator can substantially outperform state-of-the-art verbalized confidence in both in-domain and out-of-domain evaluations, as shown in Figure~\ref{fig:header}. We find that estimator overfitting can bias comparisons in favor of verbalized confidence, as continued optimization can reduce training error while degrading confidence estimation on unseen data. These results establish estimators trained with overfitting control as a strong confidence estimation paradigm that deserves further attention.

This finding also highlights the complementary strengths and limitations of estimator- and verbalization-based approaches. Hidden features within model representations are essential for accurate confidence estimation, yet they cannot be easily translated into natural language~\citep{kumaran2026llms}. At the same time, decoupling the estimator from the main policy limits its ability to fully utilize the advanced reasoning capabilities of contemporary LLMs, which have proven effective for confidence estimation and various other tasks. This raises a central question: \textit{Can we harness the advantages of both paradigms to achieve more reliable confidence estimation?}

To answer this, we propose \textbf{I}terative \textbf{Po}licy-\textbf{E}stimator \textbf{T}raining (\textbf{IPoET}) for confidence estimation. It alternates between two stages:
(i) optimizing the policy with an objective that incorporates estimator feedback while preserving task accuracy, and
(ii) updating the estimator to leverage hidden states encoding reasoning traces from the current policy.
This process enables policy reasoning to provide richer confidence signals that the estimator can extract from its hidden representations, while keeping the estimator aligned with the evolving policy distribution. Consequently, this co-adaptation yields stronger confidence estimation than isolated optimization.

We evaluate IPoET across multiple mathematical reasoning and question-answering datasets using different backbone models. Compared to both estimator- and verbalization-based approaches, our method consistently achieves superior or competitive performance on confidence estimation metrics while preserving task performance. Furthermore, IPoET generalizes effectively to out-of-domain tasks. We also investigate algorithmic variants, such as joint training, to provide deeper insights into the underlying effectiveness of our method.

Our contributions are summarized as follows:
\begin{itemize}[leftmargin=1.0em]
\item We revisit the comparison between estimator- and verbalization-based paradigms, showing that estimators can substantially outperform verbalized confidence and revealing estimator overfitting as an important factor underlying prior discrepancies.
\item We propose IPoET, an iterative training framework that harnesses the complementary advantages of LLM verbalized reasoning and rich hidden features for confidence estimation.
\item Across diverse datasets and backbone models, IPoET outperforms baselines on in-domain confidence estimation while achieving superior or comparable out-of-domain performance.
\end{itemize}

\section{Empirical Study}
\label{sec:empirical-study}

In this section, we revisit two confidence estimation paradigms: estimator-based and verbalization-based. We begin by highlighting their implementations in Section~\ref{sec:preliminaries}, before examining how estimator training dynamics affect their relative performance in Section~\ref{sec:method-comparison}.

\subsection{Preliminaries}
\label{sec:preliminaries}

\textbf{\textit{Task Definition}}: Given a prompt $x$ from a dataset $\mathcal{D}$, an LLM $\pi_\theta$ generates a response $y$ consisting of an optional reasoning trace $\tau$ and a final answer $a$. We define the correctness label as $z=\mathbb{I}[a\equiv a^*]$, where $a^*$ is the ground-truth answer. Confidence estimation aims to assign a score $c\in[0,1]$ representing the likelihood that $a$ is correct.

\paragraph{Estimator-based approaches.}
These approaches utilize an independent confidence estimator $s_\phi$. While prior work typically employs either a lightweight probing head or an entire model as the estimator, we follow \citet{malladi2023fine} and adopt the latter for its superior performance~\citep{ni2025annotation}. Specifically, $s_\phi$ processes the input $x$ and response $y$ to predict a confidence score $c$ by applying a linear head to the hidden state of the last non-padding token, denoted as $h$: $c=s_\phi(x,y)=\operatorname{clip}(\mathbf{w}^\top h+b,0,1)$, where $\mathbf{w}$ and $b$ denote the weights and bias.
We train the estimator on collected rollouts \(\mathcal{D}_{\mathrm{est}}=\{(x,y,z)\}\) using a mean squared error (MSE) loss:
\begin{equation}
\mathcal{L}_{\mathrm{est}}
=
\mathbb{E}_{(x,y,z)\sim \mathcal{D}_{\mathrm{est}}}
\left[
(s_\phi(x,y)-z)^2
\right].
\label{eq:estimator_loss}
\end{equation}
Previous studies show that critical features for confidence estimation are distributed across intermediate layers~\citep{azaria2023internal,subramani2025mice}.
Through end-to-end gradient optimization, the final hidden state learns to distill and consolidate these confidence signals scattered throughout the network.

\paragraph{Verbalization-based approaches.}
These approaches leverage the strong capabilities of recent LLMs to generate an answer $a$ alongside verbalized confidence $c$ via a natural language reasoning trace $\tau$. To jointly optimize both variables, this paradigm typically employs RL to search for reasoning traces $\tau$ that provide a more rigorous rationale for inferring a precise confidence $c$. Following RLCR~\citep{damani2025beyond}, we augment the correctness reward in conventional RLVR~\citep{shao2024deepseekmath} with a Brier-style term over the verbalized confidence:
\begin{equation}
\mathcal{J}_{\mathrm{verb}}(\theta)
=
\mathbb{E}_{x\sim\mathcal{D},\,(y,c)\sim\pi_\theta(\cdot|x)}
\left[
z-(c-z)^2
\right].
\end{equation}
Recent studies have claimed that RL-trained verbalized confidence outperforms independently trained confidence estimators~\citep{ma2026decoupling, zhang2026confidence}.
For example, in comparisons with their respective estimator baselines, RLCR highlights calibration gains, particularly under distribution shift~\citep{damani2025beyond}, while Rewarding Doubt emphasizes improved discrimination and generalization~\citep{bani2025rewarding}.

\subsection{Estimator-Based vs. Verbalization-Based}
\label{sec:method-comparison}
We next revisit the comparison between estimator-based and verbalization-based confidence estimation. We hypothesize that estimator overfitting may explain the previously claimed advantage of verbalized confidence. An evaluation at a single training endpoint may obscure this effect, as it does not reveal how training and validation performance evolve during optimization. To investigate this possibility, we examine checkpoints throughout full-parameter tuning, testing whether continued optimization yields sustained improvements or reverses earlier gains in generalization. We use the HotpotQA training set and track Brier scores on a 5K training subset, a held-out in-domain validation set, and the out-of-domain benchmarks described in Section~\ref{sec:experiment-setup}. 

As shown in Figure~\ref{fig:training-sensitivity}, the training Brier score decreases continuously, while the in-domain validation score first decreases and then rises. The out-of-domain score exhibits a more pronounced deterioration, increasing from an early minimum of 0.184 to 0.258 with extended training. This divergence provides evidence of overfitting. For subsequent estimator training, we adopt the \(1\times\) configuration without additional checkpoint selection, as it achieves in-domain validation performance close to that of \(2\times\) with half the budget. Consistent with this analysis, the full-scale experiments in Section~\ref{sec:main-results} further demonstrate strong estimator performance with overfitting control.

Our comparison shows that estimator-based confidence estimation can substantially outperform RL-trained verbalized confidence. As shown in Figure~\ref{fig:header}, the confidence estimator improves AUROC by 0.161 in-domain and 0.078 out-of-domain, while reducing Brier score by 0.049 and 0.020, respectively. These results challenge the view that verbalized confidence is inherently stronger than hidden-state estimation, highlighting the value of representation-level confidence evidence.

Beyond clarifying the relative performance of the two paradigms, this finding also leads us to examine their complementary capabilities. While estimators extract features from hidden states, RL-based verbalized methods connect confidence expression with the reasoning process~\citep{tao2024trust}. These distinct advantages suggest a natural opportunity to combine both approaches, which we explore through iterative updates of generation and confidence estimation in the next section.

\begin{figure}[t]
  \centering
  \includegraphics[width=\textwidth]{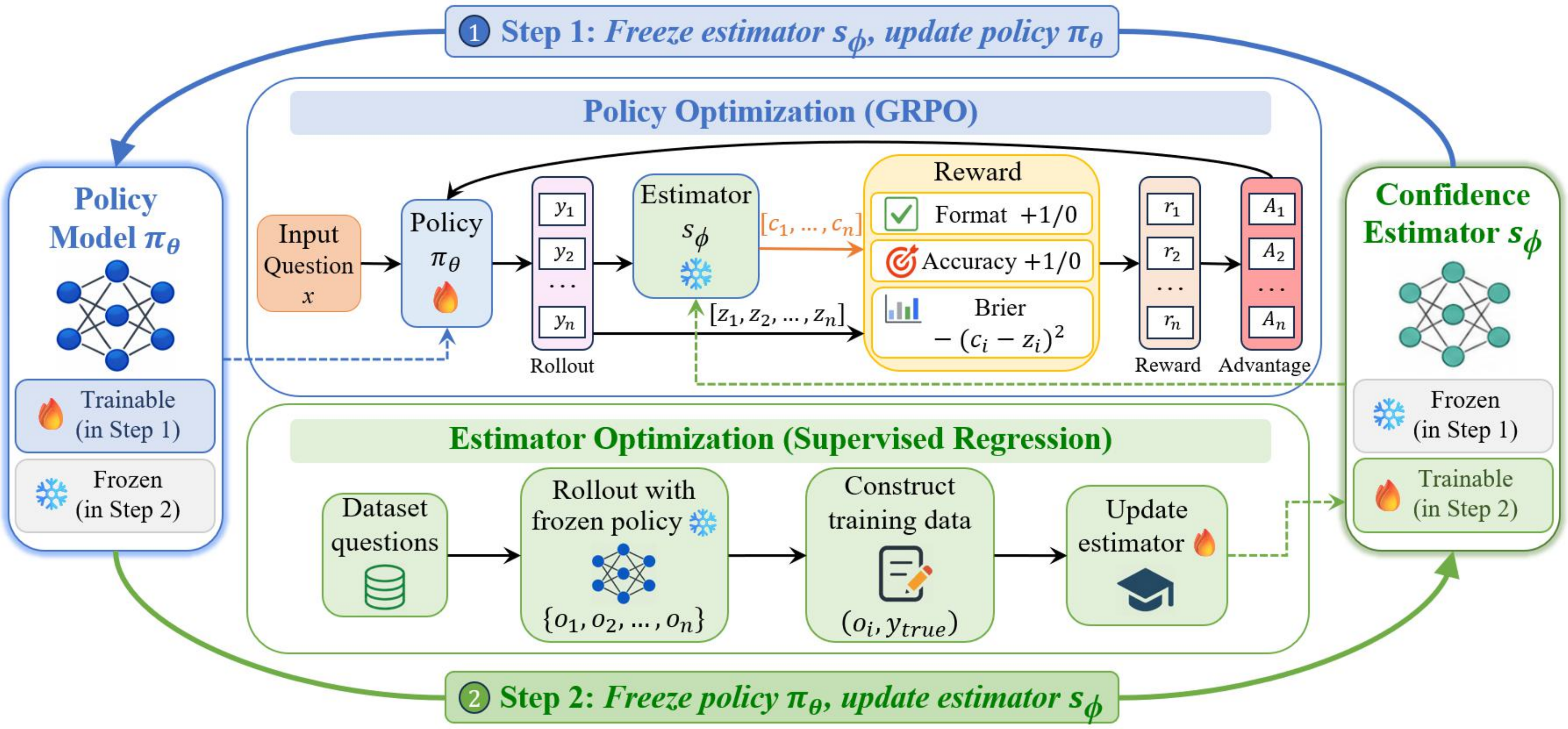}
  \caption{Overview of IPoET. The framework alternates between policy and estimator optimization. \textit{In the policy-update stage}, the confidence estimator is frozen and provides confidence estimates to form an estimator-derived term, combined with format and accuracy rewards for GRPO. \textit{In the estimator-update stage}, the policy is frozen to generate correctness-labeled rollouts for updating the estimator via supervised regression. Across iterations, the policy refines reasoning-oriented generation while the estimator adapts to the evolving policy distribution.}
  \label{fig:method}
\end{figure}

\section{Method}
\label{sec:method}
We propose IPoET, an iterative framework that connects reasoning-oriented policy training with hidden-state confidence estimation, as illustrated in Figure~\ref{fig:method}. To better transfer confidence signals into policy learning, we adopt reinforcement learning as the optimization mechanism, enabling estimator-derived reward to directly shape the model's generated reasoning. We describe the training procedure in Section~\ref{sec:iterative-training}, and then define the estimator-derived reward in Section~\ref{sec:reward-function}.

\subsection{Iterative Policy-Estimator Training}
\label{sec:iterative-training}
Let \(\pi_{\theta_0}\) denote the base model. We first apply standard RLVR~\citep{lambert2024tulu} to obtain an initial policy \(\pi_{\theta_1}\). This warm-up stage provides a stable output format for answer extraction and correctness verification. Rollouts from \(\pi_{\theta_1}\) are then used to train the confidence estimator \(s_{\phi_1}\) with the supervised regression objective in Eq.~\ref{eq:estimator_loss}.

After initialization, IPoET proceeds in alternating rounds. At iteration \(t\), it performs two steps. \textbf{Step 1: \textit{Freeze estimator \(s_{\phi_t}\), update policy \(\pi_{\theta_t}\)}.} The estimator \(s_{\phi_t}\) is held fixed while the policy \(\pi_{\theta_t}\) is updated to \(\pi_{\theta_{t+1}}\) with GRPO~\citep{shao2024deepseekmath} using a reward that combines format, accuracy, and an estimator-derived confidence term, as detailed in Section~\ref{sec:reward-function}. This step incorporates representation-level confidence feedback into policy learning, encouraging the policy to generate reasoning that makes answer correctness easier to infer from the estimator's hidden representations. \textbf{Step 2: \textit{Freeze policy \(\pi_{\theta_{t+1}}\), update estimator \(s_{\phi_t}\)}.} Since the updated policy may induce a different response distribution, we sample fresh rollouts from \(\pi_{\theta_{t+1}}\), assign correctness labels using task-specific answer verification, and update the estimator to \(s_{\phi_{t+1}}\) with the same regression objective in Eq.~\ref{eq:estimator_loss}. Across estimator updates, we use the fixed training budget established in Section~\ref{sec:method-comparison} to reduce overfitting and maintain estimator reliability on the evolving policy distribution.

Rather than relying on a static estimator, IPoET alternates policy improvement with estimator refinement on current-policy rollouts, allowing the policy and estimator to co-adapt across rounds. The estimator serves both as a confidence predictor and as a source of feedback for reasoning generation.

\subsection{Reward Function}
\label{sec:reward-function}
Following the notation in Section~\ref{sec:preliminaries}, IPoET optimizes the policy using a reward with three components:
\begin{equation}
R_{\mathrm{IPoET}}(x,y,z;\phi) = R_{\mathrm{fmt}}(y) + R_{\mathrm{acc}}(z) + R_{\mathrm{est}}(x,y,z;\phi).
\end{equation}
Here, \(R_{\mathrm{fmt}}\) maintains the required output structure, and \(R_{\mathrm{acc}}(z)=z\) rewards responses whose final answers match the ground-truth answer.

The key component is the estimator-derived reward \(R_{\mathrm{est}}\), which introduces hidden-state confidence estimation into policy training. For each rollout, the current estimator produces a confidence score from its internal representation. IPoET converts this representation-level estimate into a policy reward through a negative Brier-style objective:
\begin{equation}
R_{\mathrm{est}}(x,y,z;\phi)
=
-
\left(s_\phi(x,y)-z\right)^2 .
\end{equation}
This term gives higher reward when estimated confidence matches correctness, thereby discouraging overconfidence and underconfidence. Thus, the estimator directly guides policy optimization using hidden-state confidence estimates, without requiring the policy to verbalize a confidence score.

\section{Experiments}
\label{sec:experiments}

\subsection{Experimental Setup}
\label{sec:experiment-setup}

\paragraph{Datasets.} We focus on tasks with well-defined correctness criteria to support reliable confidence evaluation across diverse reasoning and knowledge settings. Following the experimental setup of RLCR~\citep{damani2025beyond}, we utilize the processed HotpotQA and Big-Math datasets as our training data, derived respectively from the modified HotpotQA distractor data~\citep{yang2018hotpotqa} and the filtered Big-Math problems~\citep{albalak2025big}. For evaluation, we consider benchmarks that reflect different sources of uncertainty in confidence estimation. They fall into three categories: (i) factual question answering, including HotpotQA, TriviaQA~\citep{joshi2017triviaqa}, and SimpleQA~\citep{wei2024measuring}; (ii) commonsense and expert-level knowledge reasoning, including CommonsenseQA~\citep{talmor2019commonsenseqa} and GPQA~\citep{rein2023gpqa}; and (iii) mathematical reasoning, including MATH-500~\citep{hendrycks2021measuring}, GSM8K~\citep{cobbe2021training}, and Big-Math.

\paragraph{Evaluation Metrics.} We evaluate each method along two dimensions (details in Appendix~\ref{sec:metric-details}). For task performance, we report Accuracy, computed using dataset-specific evaluation procedures, including exact match, \textit{math\_verify}, and LLM-as-a-judge~\citep{zheng2023judging} evaluation. Given the short and objective answer formats of these benchmarks, we use automated evaluation without additional human annotation. For confidence estimation, we report Area Under the Receiver Operating Characteristic Curve (AUROC)~\citep{bradley1997use}, Brier score~\citep{glenn1950verification}, and Expected Calibration Error (ECE)~\citep{guo2017calibration}. AUROC evaluates how confidence scores order correct and incorrect predictions, reflecting the ranking quality of confidence estimates. Brier score averages the squared error between the predicted confidence and the binary correctness label, capturing how accurate each confidence estimate is at the instance level. ECE measures the mismatch between predicted confidence and empirical accuracy after grouping predictions into confidence bins.

\paragraph{Baselines.} We compare with the initial model and the following representative baselines, focusing on single-pass confidence estimation without additional inference-time sampling. \textbf{RLVR} optimizes the policy with a binary answer reward~\citep{shao2024deepseekmath} and verbalizes confidence only during evaluation. \textbf{Answer-Prob} averages the generation probabilities of all tokens in the final answer~\citep{jiang2021can}, while \textbf{P(True)} prompts the model to judge whether its own answer is correct and uses the probability of the positive label as confidence~\citep{kadavath2022language}. \textbf{RLVR + Estimator} trains a confidence estimator on RLVR-generated outputs and correctness labels, following the training protocol in Section~\ref{sec:empirical-study}. \textbf{RLCR} trains the model to generate both an answer and verbalized confidence by optimizing a reward that combines answer correctness with a negative Brier score~\citep{damani2025beyond}. \textbf{RLCR (Step-aligned)} uses the RLCR objective with the same number of training steps as IPoET, serving as a control for the effect of additional training. \textbf{RLCR + Estimator} applies a confidence estimator to RLCR-generated outputs. We also evaluate commercial frontier models and observe limitations in their verbalized confidence estimation; see Appendix~\ref{sec:commercial-model-comparison}.

\paragraph{Implementation Details.} Our method is built on the verl framework~\citep{sheng2025hybridflow}, with GRPO as the RL algorithm~\citep{shao2024deepseekmath}. We use Qwen3-8B-Base~\citep{yang2025qwen3} as the main backbone and evaluate Llama-3.1-8B-Instruct~\citep{grattafiori2024llama} to assess generality. For a fair comparison, we reimplement the RL-based baselines in the same training pipeline while preserving their objectives and core design. All methods use the same training data within each setting. For each trainable baseline, we report results from the checkpoint with the best held-out in-domain validation performance. Following common practice in RL training for reasoning models~\citep{guo2025deepseek, hu2026open}, we initialize the policy from the backbone model and train it without KL regularization. Training details, including the prompt template, are provided in Appendix~\ref{sec:training-details}.

\begin{table*}[t]
\caption{Main results for models trained on HotpotQA and Big-Math. We report accuracy and confidence-estimation metrics, including AUROC, Brier score, and ECE, for Qwen and Llama backbones. For models trained on HotpotQA, we report results on HotpotQA and a six-dataset OOD average.
For models trained on Big-Math, we report a Math average over MATH-500, GSM8K, and Big-Math, together with a five-dataset OOD average. Best and second-best results for all metrics under each backbone are marked in \textbf{bold} and \underline{underlined}, respectively.}
\label{tab:main-results}
\begin{center}
\small
\setlength{\tabcolsep}{4.2pt}
\setlength{\dashlinedash}{2pt}
\setlength{\dashlinegap}{1.2pt}
\resizebox{0.95\textwidth}{!}{
\begin{tabular}{llcccccccc}
\toprule
& \multicolumn{4}{c}{\textbf{HotpotQA}}
& \multicolumn{4}{c}{\textbf{OOD Averaged}} \\
\cmidrule(lr){2-5} \cmidrule(lr){6-9}
& Acc.\,$\uparrow$ & AUROC\,$\uparrow$ & Brier\,$\downarrow$ & ECE\,$\downarrow$
& Acc.\,$\uparrow$ & AUROC\,$\uparrow$ & Brier\,$\downarrow$ & ECE\,$\downarrow$ \\
\midrule

Qwen3-8B-Base
& 51.6\% & 0.555 & 0.363 & 0.343
& 56.5\% & 0.556 & 0.359 & 0.345 \\

\quad $\llcorner$ RLVR
& 62.2\% & 0.523 & 0.376 & 0.376
& \textbf{62.3\%} & 0.518 & 0.357 & 0.358 \\

\quad $\llcorner$  Answer-Prob
& 62.2\% & 0.661 & 0.359 & 0.360
& \textbf{62.3\%} & 0.548 & 0.340 & 0.327 \\

\quad $\llcorner$  P(True)
& 62.2\% & 0.543 & 0.308 & 0.260
& \textbf{62.3\%} & 0.588 & 0.304 & 0.314 \\

\quad $\llcorner$  RLVR + Estimator
& 62.2\% & \underline{0.757} & \underline{0.195} & \underline{0.087}
& \textbf{62.3\%} & \underline{0.724} & \underline{0.184} & 0.184 \\

\quad $\llcorner$  RLCR
& 61.0\% & 0.596 & 0.244 & 0.100
& 61.4\% & 0.646 & 0.204 & \underline{0.171} \\

\quad $\llcorner$  RLCR (Step-aligned)
& \textbf{65.3\%} & 0.606 & 0.263 & 0.205
& \underline{61.6\%} & 0.690 & 0.210 & 0.188 \\

\quad $\llcorner$  RLCR + Estimator
& 61.0\% & 0.753 & 0.201 & \textbf{0.081}
& 61.4\% & 0.719 & 0.207 & 0.224 \\

\rowcolor{my_color}\quad $\llcorner$  IPoET (ours)
& \underline{63.3\%} & \textbf{0.800} & \textbf{0.184} & 0.107
& \textbf{62.3\%} & \textbf{0.738} & \textbf{0.170} & \textbf{0.146} \\

\hdashline
\addlinespace[0.05em]
Llama-3.1-8B-Instruct
& 45.9\% & 0.577 & 0.461 & 0.469
& 45.9\% & 0.606 & 0.413 & 0.424 \\

\quad $\llcorner$  RLVR + Estimator
& \underline{61.6\%} & \underline{0.772} & \underline{0.198} & \underline{0.110}
& \underline{51.0\%} & \textbf{0.746} & \textbf{0.180} & \underline{0.161} \\

\quad $\llcorner$  RLCR
& 59.1\% & 0.603 & 0.248 & 0.121
& 50.2\% & 0.674 & 0.217 & 0.175 \\

\rowcolor{my_color}\quad $\llcorner$ IPoET (ours)
& \textbf{64.6\%} & \textbf{0.781} & \textbf{0.188} & \textbf{0.101}
& \textbf{51.5\%} & \underline{0.737} & \underline{0.182} & \textbf{0.152} \\

\bottomrule
\end{tabular}
}

\vspace{0.55em}

\resizebox{0.95\textwidth}{!}{
\begin{tabular}{llcccccccc}
\toprule

& \multicolumn{4}{c}{\textbf{Math}}
& \multicolumn{4}{c}{\textbf{OOD Averaged}} \\
\cmidrule(lr){2-5} \cmidrule(lr){6-9}
& Acc.\,$\uparrow$ & AUROC\,$\uparrow$ & Brier\,$\downarrow$ & ECE\,$\downarrow$
& Acc.\,$\uparrow$ & AUROC\,$\uparrow$ & Brier\,$\downarrow$ & ECE\,$\downarrow$ \\
\midrule

Qwen3-8B-Base
& 62.8\% & 0.568 & 0.313 & 0.290
& 51.9\% & 0.562 & 0.389 & 0.382 \\

\quad $\llcorner$ RLVR
& \textbf{74.4\%} & 0.517 & 0.255 & 0.255
& \textbf{52.9\%} & 0.569 & 0.417 & 0.424 \\

\quad $\llcorner$ Answer-Prob
& \textbf{74.4\%} & 0.709 & 0.235 & 0.239
& \textbf{52.9\%} & 0.576 & 0.420 & 0.424 \\

\quad $\llcorner$ P(True)
& \textbf{74.4\%} & 0.490 & 0.366 & 0.380
& \textbf{52.9\%} & 0.645 & 0.275 & 0.237 \\

\quad $\llcorner$ RLVR + Estimator
& \textbf{74.4\%} & \underline{0.862} & \underline{0.107} & \textbf{0.037}
& \textbf{52.9\%} & \underline{0.684} & \underline{0.217} & \underline{0.197} \\


\quad $\llcorner$ RLCR
& 73.7\% & 0.625 & 0.184 & 0.115
& \underline{52.8\%} & 0.600 & 0.282 & 0.272 \\

\quad $\llcorner$ RLCR (Step-aligned)
& 73.8\% & 0.633 & 0.181 & 0.120
& 41.7\% & 0.536 & 0.361 & 0.347 \\

\quad $\llcorner$ RLCR + Estimator
& 73.7\% & 0.828 & 0.126 & 0.068
& 52.8\% & 0.662 & 0.249 & 0.249 \\

\rowcolor{my_color}\quad $\llcorner$ IPoET (ours)
& \underline{74.2\%} & \textbf{0.878} & \textbf{0.100} & \underline{0.040}
& 50.9\% & \textbf{0.698} & \textbf{0.206} & \textbf{0.183} \\

\hdashline
\addlinespace[0.05em]

Llama-3.1-8B-Instruct
& 42.7\% & 0.591 & 0.501 & 0.512
& 43.7\% & 0.611 & 0.410 & 0.423 \\

\quad $\llcorner$ RLVR + Estimator
& \underline{53.8\%} & \underline{0.789} & \underline{0.167} & \underline{0.085}
& \textbf{45.3\%} & \underline{0.616} & 0.235 & 0.216 \\

\quad $\llcorner$ RLCR
& 50.6\% & 0.570 & 0.246 & 0.165
& 44.6\% & \underline{0.616} & \textbf{0.219} & \textbf{0.179} \\

\rowcolor{my_color}\quad $\llcorner$ IPoET (ours)
& \textbf{56.5\%} & \textbf{0.824} & \textbf{0.152} & \textbf{0.061}
& \underline{45.2\%} & \textbf{0.622} & \underline{0.226} & \underline{0.191} \\

\bottomrule
\end{tabular}
}
\end{center}
\end{table*}

\subsection{Main Results}
\label{sec:main-results}

\paragraph{IPoET achieves strong overall in-domain performance.} Table~\ref{tab:main-results} shows that IPoET achieves the highest AUROC and the lowest Brier score across all four in-domain settings, covering both datasets and backbone models. Compared with RLVR + Estimator, IPoET achieves AUROC gains and Brier score reductions of up to 0.043 and 0.015, respectively, highlighting its strengths in confidence discrimination and calibration. A paired bootstrap test on the Qwen3-8B-Base HotpotQA setting shows statistically significant improvements over RLVR + Estimator in both AUROC ($p=0.0006$) and Brier score ($p=0.0317$). IPoET also achieves the highest task accuracy and the lowest ECE in both Llama settings, with competitive results on both measures for Qwen3. Together, these results support the value of iterative policy–estimator interaction for improving confidence estimation. The baseline comparisons further corroborate our empirical study: with validation-based checkpoint selection, RLVR + Estimator outperforms RLCR in AUROC, Brier score, and ECE across all four in-domain settings, suggesting that hidden-state estimators can exploit correctness-related information not fully captured by explicit self-reports. Despite the additional verbalized confidence analysis in RLCR-generated outputs, RLCR + Estimator is worse than RLVR + Estimator on most metrics. One possible explanation is that explicit self-assessment may dilute correctness-related signals in the original reasoning trace and introduce extra noise into the estimator's hidden representations.

\paragraph{IPoET achieves competitive out-of-domain performance.} On benchmarks outside the training domain, IPoET ranks first or second on all confidence-estimation metrics. Under both HotpotQA and Big-Math training, it achieves the highest out-of-domain AUROC and lowest Brier score and ECE on Qwen. On Llama, IPoET achieves the lowest ECE under HotpotQA training and the highest AUROC under Big-Math training, while remaining close to the best baseline results on the other metrics. These results suggest that the benefits of IPoET extend beyond the training domain, although improvements over existing methods vary across evaluation settings.

\paragraph{IPoET's gains stem from iterative policy-estimator coupling rather than additional training alone.} RLCR (Step-aligned) extends RLCR to the same number of training updates as IPoET, providing a controlled comparison for assessing continued optimization. Additional RLCR training does not yield consistent gains and can degrade out-of-domain performance under Big-Math training. This contrast suggests that the benefit of additional optimization depends on how the policy and confidence estimator are jointly refined, rather than on the number of updates alone. IPoET therefore makes more effective use of the extra training budget for confidence estimation.

\begin{table}[t]
\caption{Comparison between joint and iterative training on HotpotQA. Iterative training improves accuracy, AUROC, and Brier score in both in-domain and out-of-domain settings. Values in parentheses denote changes relative to joint training.}
\label{tab:joint-vs-iterative}
\begin{center}
\small
\setlength{\tabcolsep}{5pt}
\begin{tabular}{ccccc}
\toprule
Setting & Strategy & Accuracy\,$\uparrow$ & AUROC\,$\uparrow$ & Brier\,$\downarrow$ \\
\midrule
\multirow{2}{*}{In-domain}
& Joint & 60.5\% & 0.773 & 0.209 \\
& Iterative & \textbf{63.3\%} {\scriptsize (+2.8 pp)}
            & \textbf{0.800} {\scriptsize (+0.027)}
            & \textbf{0.184} {\scriptsize (-0.025)} \\
\midrule
\multirow{2}{*}{Out-of-domain}
& Joint & 58.4\% & 0.737 & 0.180 \\
& Iterative & \textbf{62.3\%} {\scriptsize (+3.9 pp)}
            & \textbf{0.738} {\scriptsize (+0.001)}
            & \textbf{0.170} {\scriptsize (-0.010)} \\
\bottomrule
\end{tabular}
\end{center}
\end{table}

\subsection{Ablation and Analysis}

\paragraph{Effect of Estimator Feedback and Iterative Updates.}
We first analyze how the two stages of IPoET contribute to its performance. As shown in Figure~\ref{fig:step-wise-effects}, with the estimator \(s_{\phi_1}\) fixed, updating the policy from \(\pi_{\theta_1}\) to \(\pi_{\theta_2}\) increases AUROC and reduces Brier score in both in-domain and out-of-domain evaluations. ECE also decreases in both settings, with the in-domain score falling from 0.087 to 0.055. Alongside these gains, in-domain accuracy increases by 1.1 percentage points, while out-of-domain accuracy is preserved. The first-stage results are consistent with the intended role of estimator feedback: rather than treating confidence estimation solely as a prediction task over given responses, IPoET uses the fixed estimator’s prediction errors to shape response generation, encouraging outputs that better support reliable correctness assessment. Moreover, computing this reward from estimator predictions rather than verbalized confidence allows IPoET to guide policy optimization with the more accurate confidence estimates observed in our empirical study. With the policy held fixed, the subsequent estimator update further improves confidence quality, especially out of domain. These step-wise gains indicate complementary roles of the two stages: policy optimization incorporates estimator feedback into generation behavior, while estimator refinement adapts confidence estimation to the updated policy distribution. Together, these results support the iterative interaction between policy and estimator underlying IPoET.

\noindent
\begin{minipage}[t]{0.47\textwidth}
  \vspace{0pt}
  \centering
  \includegraphics[width=\linewidth]{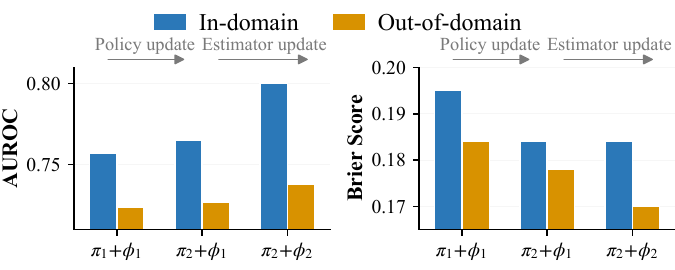}
  \captionof{figure}{
    Step-wise effects of IPoET, showing that both policy and estimator
    updates progressively improve confidence estimation.
  }
  \label{fig:step-wise-effects}
\end{minipage}
\hfill
\begin{minipage}[t]{0.52\textwidth}
  \vspace{0pt}
  \centering
  \vspace{\baselineskip}

  \captionof{table}{
    Effect of alternating-update granularity under Big-Math training
    with the same update budget and data allocation.
    Best results are shown in \textbf{bold}.
  }
  \label{tab:update-granularity}

  \small
  \setlength{\tabcolsep}{3.5pt}
  \begin{tabular}{ccccc}
    \toprule
    \multirow{2}{*}{
      \raisebox{-2.5ex}{
        \shortstack[c]{\textbf{Iteration}\\[-0.5pt]\textbf{Setting}}
      }
    }
    & \multicolumn{2}{c}{\textbf{In-domain}}
    & \multicolumn{2}{c}{\textbf{Out-of-domain}} \\
    \cmidrule(lr){2-3}
    \cmidrule(lr){4-5}
    & AUROC\,$\uparrow$
    & Brier\,$\downarrow$
    & AUROC\,$\uparrow$
    & Brier\,$\downarrow$ \\
    \midrule
    $2{\times}200$
    & \textbf{0.878}
    & \textbf{0.100}
    & \textbf{0.698}
    & 0.206 \\
    $4{\times}100$
    & 0.873
    & 0.102
    & 0.696
    & \textbf{0.198} \\
    \bottomrule
  \end{tabular}
\end{minipage}

\paragraph{Joint vs. Iterative Training.}
We next examine how estimator feedback should be incorporated into policy optimization by comparing two training designs developed in our work. Table~\ref{tab:joint-vs-iterative} shows that iterative training yields higher AUROC and lower Brier score than joint training across both evaluation settings, indicating better confidence estimation. More importantly, joint optimization degrades answer accuracy relative to the starting policy, whereas iterative training improves task performance, further supporting the design of IPoET. In joint training, the estimator must learn from a changing policy distribution while the policy is guided by an under-adapted estimator, potentially introducing interference between policy learning and estimator fitting. IPoET instead separates these roles across rounds: the estimator adapts to current policy outputs before being used to construct the confidence term in the next policy update, enabling more stable confidence-aware training.

\paragraph{Inference Efficiency and Estimator Overhead.}
IPoET introduces an additional confidence estimator with the same backbone architecture as the policy. To assess its practical computational cost, we measure total end-to-end inference time on 1,000 HotpotQA examples. IPoET requires 500.38~s, compared with 554.60~s for RLCR, representing a 9.8\% reduction. Although the estimator increases the parameter count, it scores each response in a single forward pass, whereas RLCR performs additional autoregressive generation for confidence-related reasoning. Overall, this trade-off allows IPoET to remain inference-efficient in practice despite the additional model component.

\paragraph{Granularity of Alternating Updates.}
A natural question is whether IPoET benefits from more frequent alternation between policy and estimator updates. To isolate this factor, we compare two schedules with the same update budget and data allocation: $2{\times}200$ uses two longer update phases, while $4{\times}100$ splits each corresponding phase into two shorter ones. Table~\ref{tab:update-granularity} shows that more frequent alternation does not consistently improve confidence estimation. The $2{\times}200$ schedule performs better in three of the four comparisons, including AUROC in both in-domain and out-of-domain evaluations and in-domain Brier score. These results suggest that sustained optimization within each phase better supports policy–estimator adaptation than switching more often. When each phase becomes too short, the policy may not fully absorb the estimator feedback before the estimator is updated again, yielding smaller gains in confidence estimation.

\begin{figure}[h]
  \centering
  \includegraphics[width=\textwidth]{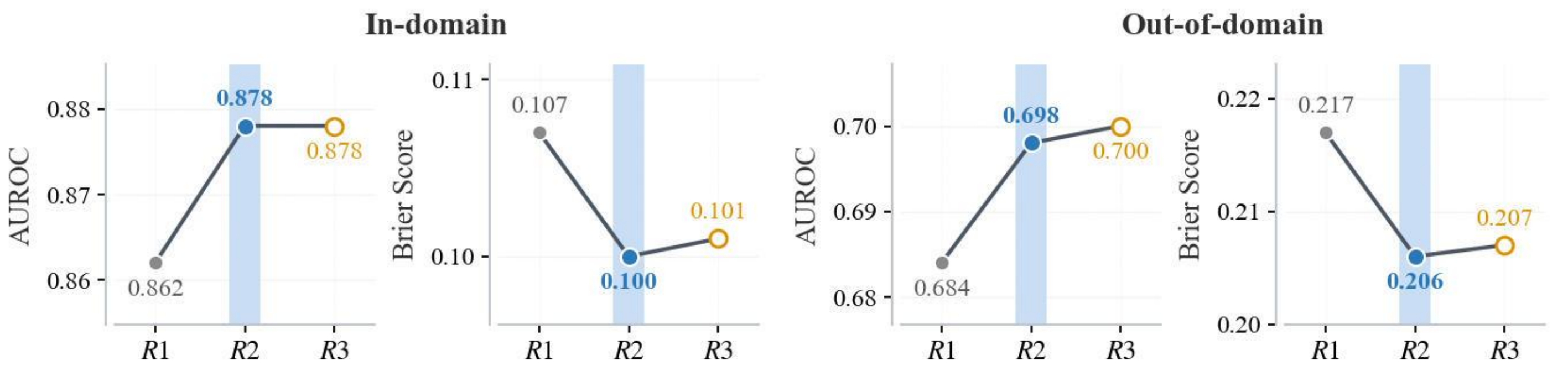}
  \caption{Round-wise comparison of in-domain and out-of-domain AUROC and Brier score under Big-Math training. Blue shading marks the configuration used in our main experiments (Round 2).}
  \label{fig:number-of-iter}
\end{figure}

\paragraph{Impact of Iteration Rounds.}
We finally examine whether extending IPoET to additional rounds further improves confidence estimation. The trend in Figure~\ref{fig:number-of-iter} shows that the transition from Round 1 to Round 2 substantially improves both AUROC and Brier score on in-domain and out-of-domain evaluations, confirming the benefit of the first complete policy–estimator refinement cycle. Round 3 brings only a marginal improvement in out-of-domain AUROC, while in-domain AUROC remains unchanged and Brier score slightly increases in both settings. This pattern suggests diminishing returns from further alternation in this setting. We therefore adopt the Round 2 configuration for our main experiments, balancing confidence discrimination and calibration quality while avoiding an additional training round.

\section{Related Work}
\paragraph{Confidence Estimation from Implicit Information.}
These methods estimate confidence from implicit information produced during inference or internal computation. One line of work studies model-internal representations for confidence estimation. LLM hidden activations have been shown to encode factuality and truthfulness-related information~\citep{chen2024inside}, motivating lightweight probes or classifiers trained on internal states~\citep{azaria2023internal, servedio2025hidden}. Subsequent work extends this view to reasoning settings, including hidden-state verification of reasoning steps and perturbation-based confidence probing~\citep{zhang2025reasoning, khanmohammadi2025calibrating}. Recent work exploits internal representations by aggregating hidden-state information across multiple layers during self-evaluation for uncertainty estimation~\citep{xiao2026enhancing}. Another line derives implicit confidence from output probabilities or self-evaluation signals: Answer-Prob uses token likelihoods or answer probabilities as confidence proxies~\citep{jiang2021can, gupta2024language}, while P(True) asks the model to judge whether its own answer is correct and uses the probability of a positive judgment~\citep{kadavath2022language}. However, these signals are usually used to score fixed outputs, whereas IPoET converts internal-representation estimates into feedback for policy training.

\paragraph{Verbalized Confidence Estimation.}
Verbalized methods require the model to express confidence explicitly through natural language. Prompting-based approaches elicit numerical or linguistic confidence alongside generated answers or reasoning~\citep{xiong2024can, yang2024verbalized}. Training-based approaches further teach models to produce uncertainty rationales or calibrated confidence scores, such as self-reflective rationales in SaySelf~\citep{xu2024sayself}, listener-aware confidence markers in LACIE~\citep{stengel2024lacie}, and calibration-aware on-policy distillation~\citep{zhang2026illusion}. Studies of reasoning models show that chain-of-thought generation can improve confidence expression~\citep{yoon2026reasoning}, motivating recent RL-based methods that optimize verbalized confidence with calibration-oriented rewards, such as Brier-style or logarithmic scoring objectives~\citep{damani2025beyond, bani2025rewarding}. However, expressed confidence remains sensitive to prompt formulation, answer-dependent self-assessment, and placement of confidence statements relative to the answer~\citep{xia2025influences, seo2025advice, guo2026llms, li2026orce}. IPoET avoids relying on such explicit confidence expression by deriving confidence from an internal-representation estimator, while still benefiting from reasoning-oriented generation.

\section{Conclusion}
This work revisits two major paradigms for LLM confidence estimation: estimator-based and verbalization-based methods. We demonstrate that estimators can substantially outperform verbalized confidence when overfitting is controlled, partly explaining the differing conclusions in prior comparisons. Nevertheless, traditional estimators fail to utilize the strong reasoning capabilities of contemporary LLMs. To combine the strengths of both paradigms, we propose IPoET (Iterative Policy-Estimator Training). This framework enables the estimator to extract richer features from its hidden states using policy-generated reasoning traces, while the policy learns to produce more informative content for the estimator. Compared with the baselines, IPoET shows improved in-domain confidence estimation, comparable or better out-of-domain performance, and competitive task accuracy across multiple reasoning and question-answering datasets. These results highlight the value of internal representations for reliable confidence estimation and suggest that iterative policy-estimator training is a promising direction for building more trustworthy reasoning models.

\subsection*{AI use statement}

We used generative AI tools to improve the wording, grammar, and fluency of the manuscript and refine the presentation of scientific figures. These tools were not used to implement the proposed method, generate datasets, or prove mathematical claims. All reported numerical results and the data underlying the figures were obtained from experiments conducted by the authors. We reviewed all AI-assisted work. Specifically, we checked textual revisions for technical accuracy and consistency with our intended meaning. The revised figures were also verified against the underlying experimental data. We take responsibility for the final content of this work, including text, claims or artifacts produced with the aid of generative AI.



\subsection*{Ethics statement}


We are committed to adhering to the ICLR Code of Ethics. Our study relies solely on publicly available benchmark datasets for mathematical reasoning and question answering, without recruiting human participants or collecting private personal data. We believe this work raises no direct ethical risks beyond standard concerns associated with LLM research and use.

\subsection*{Reproducibility statement}


To support reproducibility, we provide our source code at \url{https://github.com/xyk829/ipoet}. The training procedure and reward function are described in Sections~\ref{sec:iterative-training} and~\ref{sec:reward-function}, respectively. Our experiments use publicly available backbone models and benchmark datasets, with the experimental setup described in Section~\ref{sec:experiment-setup}. Additional training details and the prompt template are provided in Appendix~\ref{sec:training-details}, while evaluation procedures and metric definitions are detailed in Appendix~\ref{sec:metric-details}.



\bibliography{iclr2027_conference}
\bibliographystyle{iclr2027_conference}

\clearpage
\appendix

\section{Training Details}
\label{sec:training-details}
All experiments are conducted on a server equipped with 8 NVIDIA A100 GPUs, each with 80GB of memory. For both HotpotQA and Big-Math, the overall training time is approximately 15 hours.

For policy training, we use the same hyperparameters in both rounds. Specifically, we use GRPO with 8 sampled responses per prompt and a sampling temperature of 0.7. The policy learning rate is set to 1e-6 for both HotpotQA and Big-Math. In line with \citet{damani2025beyond}, we use a training batch size of 64 for HotpotQA and 72 for Big-Math. The maximum prompt and response lengths are set to 3072 and 1536 for HotpotQA, and 1024 and 4096 for Big-Math, respectively. We do not normalize GRPO advantages by their group standard deviation, following prior practice~\citep{bereket2025uncalibrated, turtel2025outcome}, thereby preserving reward-scale information for examples with large calibration errors.

For estimator training, we use a per-device batch size of 1 with gradient accumulation over 8 steps. We reduce the second-round learning rate to support stable refinement and avoid overfitting or excessive drift from the first-round estimator. Specifically, the learning rate decreases from $5 \times 10^{-6}$ to $1 \times 10^{-6}$ across the two HotpotQA rounds and from $5 \times 10^{-6}$ to $5 \times 10^{-7}$ across the two Big-Math rounds.

The prompt template used for policy training is shown in Figure~\ref{fig:training-prompt}.

\begin{figure}[h]
\begin{center}
\begin{promptbox}
A conversation between User and Assistant. The user asks a question, and the Assistant solves it. The assistant first thinks about the reasoning process in the mind and then provides the user with the answer. The reasoning process and answer are enclosed within \texttt{<reason> </reason>} and \texttt{<answer> </answer>} tags, respectively, i.e., \texttt{<reason>} reasoning process here \texttt{</reason><answer>} answer here \texttt{</answer>}.
\end{promptbox}
\end{center}
\caption{Policy training prompt template.}
\label{fig:training-prompt}
\end{figure}

\section{Details of Evaluation Metrics}
\label{sec:metric-details}
We follow the notation defined in the main text.
\paragraph{Accuracy.}
We compute accuracy using dataset-specific correctness criteria. For HotpotQA and HotpotQA-Modified, we use exact match. For GSM8K, MATH-500, and Big-Math, we use the \textit{math\_verify} library. For CommonsenseQA, GPQA, SimpleQA, and TriviaQA, we use an LLM-as-a-judge~\citep{zheng2023judging} evaluation. To keep the judge model family separate from the evaluated backbone, we use Llama-3.1-8B-Instruct for Qwen-based methods and Qwen3-8B for Llama-based methods. Following \citet{damani2025beyond}, we run the judge with temperature set to 0 and provide it with the question, the ground-truth answer, and the extracted answer. The judge is instructed to output only “YES” or “NO” according to whether the answer is correct. Since these datasets contain short and objective answers, we do not condition the judge on thinking traces, avoiding potential bias from generated rationales.

\paragraph{AUROC.}
We compute AUROC by sweeping the confidence threshold and integrating the resulting ROC curve:
\[
\mathrm{AUROC}
=
\int_{0}^{1}
\mathrm{TPR}\left(\mathrm{FPR}^{-1}(x)\right)
\,dx,
\]
where TPR and FPR denote the true positive rate and false positive rate, respectively.

\paragraph{Brier score.}
Given confidence scores $c_i$ and correctness labels $z_i$, we compute
\[
\mathrm{Brier\ score}
=
\frac{1}{N}\sum_{i=1}^{N}(c_i-z_i)^2 .
\]

\paragraph{ECE.}
We partition predictions into $M=10$ confidence bins $\{B_m\}_{m=1}^{M}$. For each bin, we compute the average correctness and average confidence:
\[
\bar{z}_m=\frac{1}{|B_m|}\sum_{i\in B_m}z_i .
\]
\[
\bar{c}_m=\frac{1}{|B_m|}\sum_{i\in B_m}c_i .
\]
The ECE is then computed as
\[
\mathrm{ECE}
=
\sum_{m=1}^{M}
\frac{|B_m|}{N}
\left|\bar{z}_m-\bar{c}_m\right| .
\]

\begin{table}[h]
\caption{Comparison between IPoET and frontier commercial models. IPoET is trained on HotpotQA for the HotpotQA results and on Big-Math for the math average over MATH-500, GSM8K, and Big-Math. Best and second-best results are marked in \textbf{bold} and \underline{underlined}, respectively.}
\label{tab:commercial-comparison}
\begin{center}
\small
\setlength{\tabcolsep}{6pt}

\textbf{(a) HotpotQA}

\vspace{0.2em}

\begin{tabular}{@{}lccc@{}}
\toprule
\textbf{Model} & \textbf{AUROC}$\uparrow$ & \textbf{Brier}$\downarrow$ & \textbf{ECE}$\downarrow$ \\
\midrule
GPT-5 mini          & 0.603 & 0.306 & 0.298 \\
Claude Haiku 4.5    & \underline{0.671} & \underline{0.279} & \underline{0.266} \\
DeepSeek-V4-Flash   & 0.537 & 0.332 & 0.328 \\
Gemini 3 Flash      & 0.509 & 0.357 & 0.357 \\
\rowcolor{my_color}
IPoET (ours)         & \textbf{0.800} & \textbf{0.184} & \textbf{0.107} \\
\bottomrule
\end{tabular}

\vspace{0.6em}

\textbf{(b) Math average}

\vspace{0.2em}

\begin{tabular}{@{}lccc@{}}
\toprule
\textbf{Model} & \textbf{AUROC}$\uparrow$ & \textbf{Brier}$\downarrow$ & \textbf{ECE}$\downarrow$ \\
\midrule
GPT-5 mini          & 0.638 & 0.189 & 0.181 \\
Claude Haiku 4.5    & 0.584 & 0.309 & 0.300 \\
DeepSeek-V4-Flash   & \underline{0.686} & \underline{0.137} & \underline{0.136} \\
Gemini 3 Flash      & 0.557 & 0.150 & 0.150 \\
\rowcolor{my_color}
IPoET (ours)         & \textbf{0.878} & \textbf{0.100} & \textbf{0.040} \\
\bottomrule
\end{tabular}
\end{center}
\end{table}

\section{Commercial Model Comparison}
\label{sec:commercial-model-comparison}
We further compare IPoET with commercial models in Table~\ref{tab:commercial-comparison}. We evaluate GPT-5 mini~\citep{gpt5}, Claude Haiku 4.5~\citep{claude4}, DeepSeek-V4-Flash~\citep{deepseek2026v4flash}, and Gemini 3 Flash~\citep{google2025gemini3flash}.\footnote{Model versions used: \path{gpt-5-mini-2025-08-07}, \path{claude-haiku-4-5-20251001}, \path{deepseek-v4-flash}, and \path{gemini-3-flash-preview-nothinking}.} For the HotpotQA results, IPoET is trained on HotpotQA; for the math results, IPoET is trained on Big-Math and evaluated using the Math average over MATH-500, GSM8K, and Big-Math. Across both settings, IPoET obtains the best results on AUROC, Brier score, and ECE among the evaluated models. Commercial models demonstrate reasonable ability to estimate confidence, but their verbalized confidence remains less reliable in these settings. These results indicate that strong general-purpose models can still struggle to align expressed confidence with correctness on reasoning tasks, highlighting the need for task-oriented training objectives that explicitly target confidence reliability.

\end{document}